\documentclass[a4paper, 10pt, conference]{ieeeconf}      % Use this line for a4 paper

\IEEEoverridecommandlockouts                              % This command is only needed if 
\usepackage{graphics} % for pdf, bitmapped graphics files
\usepackage{epsfig} % for postscript graphics files
\usepackage{amsmath} % assumes amsmath package installed
\usepackage{amssymb}  % assumes amsmath package installed

\usepackage{soul}
\usepackage{algorithm}
\usepackage{graphicx}
\usepackage{xcolor}

\usepackage{csquotes}

\usepackage{wrapfig}
\usepackage[noend]{algpseudocode}
\usepackage[tight]{subfigure}
\usepackage[nolist]{acronym}
\usepackage{placeins}
\usepackage{dsfont}

\makeatletter
\let\NAT@parse\undefined
\makeatother

\usepackage[colorlinks]{hyperref}

\hypersetup{
    colorlinks=true,
    citecolor=blue,
    linkcolor=blue,
    filecolor=magenta,      
    urlcolor=cyan,
    urlcolor=black
}

\title{\LARGE \bf
One-Shot Informed Robotic Visual Search in the Wild
}

\author{Karim Koreitem$^{1}$, Florian Shkurti$^{1}$, Travis Manderson$^{2}$, Wei-Di Chang$^{2}$,\\Juan Camilo Gamboa Higuera$^{2}$, and Gregory Dudek$^{2}$% <-this % stops a space
\thanks{$^{1}$Karim Koreitem {\tt\footnotesize k.koreitem@mail.utoronto.ca} and Florian Shkurti {\tt\footnotesize florian@cs.toronto.edu} are affiliated with the Department of Computer Science, the Robotics Institute at the University of Toronto, and Vector Institute. 
}%
\thanks{$^{2}$Travis Manderson, Wei-Di Chang, Juan Camilo Gamboa Higuera, and Gregory Dudek are affiliated with the Center for Intelligent Machines, School of Computer Science, McGill University, in Montr\'eal.
        {\tt\footnotesize \{travism, wchang, gamboa, dudek\}@cim.mcgill.ca}.}%
}

\begin{document}

\maketitle
\thispagestyle{empty}
\pagestyle{empty}

%%%%%%%%%%%%%%%%%%%%%%%%%%%%%%%%%%%%%%%%%%%%%%%%%%%%%%%%%%%%%%%%%%%%%%%%%%%%%%%%
\begin{abstract}
%
% Problem: data collection robots mainly operate through path tracking,
% without taking into account the quality of data they collect, as judged by 
% a human
%
% We propose:
%  - a Weakly supervised method for learning representations from video that 
%    facilitate similarity search with a single exemplar
%  - we encourage viewpoint invariance in the representation
%  - a navigation method that makes use of the learned similarity search 
%    method for collecting useful data from a single example in an unknown
%    environment   
%
% We evaluate:
%  - classification accuracy (similar or not) on two underwater datasets
%    where we show improvements over pretrained ImageNet embeddings
%  - we deploy the similarity model in collaborative visual search scenarios
%  - usefulness of data collected from informed navigation policy vs  
%    uninformed, both qualitatively and based on human labels 
%  

 We consider the task of underwater robot navigation for the purpose of collecting scientifically relevant video data for environmental monitoring. The majority of field robots that currently perform monitoring tasks in unstructured natural environments navigate via path-tracking a pre-specified sequence of waypoints. Although this navigation method is often necessary, it is limiting because the robot does not have a model of what the scientist deems to be relevant visual observations. Thus, the robot can neither visually search for particular types of objects, nor focus its attention on parts of the scene that might be more relevant than the pre-specified waypoints and viewpoints. In this paper we propose a method that enables informed visual navigation via a learned visual similarity operator that guides the robot's visual search towards parts of the scene that look like an exemplar image, which is given by the user as a high-level specification for data collection. We propose and evaluate a weakly supervised video representation learning method that outperforms ImageNet embeddings for similarity tasks in the underwater domain. We also demonstrate the deployment of this similarity operator during informed visual navigation in collaborative environmental monitoring scenarios, in large-scale field trials, where the robot and a human scientist collaboratively search for relevant visual content. Code: \url{https://github.com/rvl-lab-utoronto/visual_search_in_the_wild} 
 %Code accompanying this paper is available at  %Code accompanying this paper is available here\footnote[1]{Github repo: \url{https://github.com/rvl-lab-utoronto/visual_search_in_the_wild}}.
\end{abstract}

%%%%%%%%%%%%%%%%%%%%%%%%%%%%%%%%%%%%%%%%%%%%%%%%%%%%%%%%%%%%%%%%%%%%%%%%%%%%%%%%

\section{INTRODUCTION}
\label{sec:intro}
%
% Problem: data collection robots mainly operate through path tracking,
% without taking into account the quality of data they collect, as judged by 
% a human
%
% We propose:
%  - a Weakly supervised method for learning representations from video that 
%    facilitate similarity search with a single exemplar
%  - we encourage viewpoint invariance in the representation
%  - a navigation method that makes use of the learned similarity search 
%    method for collecting useful data from a single example in an unknown
%    environment   
%
% We evaluate:
%  - classification accuracy (similar or not) on two underwater datasets
%    where we show improvements over pretrained ImageNet embeddings
%  - we deploy the similarity model in collaborative visual search scenarios
%  - usefulness of data collected from informed navigation policy vs  
%    uninformed, both qualitatively and based on human labels 
%  

One of the main functions of mobile robots in the context of environmental monitoring is to collect scientifically relevant data for users who are not experts in robotics, but whose scientific disciplines -- oceanography, biology, ecology, geography, among others -- increasingly rely on automated data collection by mobile robots carrying scientific sensors~\cite{WILLIAMS2016158}. The areas and volumes these field robots are tasked to inspect are too vast to cover exhaustively, so a major challenge in retrieving relevant sensor data is to enable sensing in physical space, in a way that balances \textit{exploration}, to minimize epistemic uncertainty and infer the correct physical model,  and \textit{exploitation}, to record data that the user knows they will be interested in. There has been significant progress in addressing the exploration problem in terms of active sensing and informed path planning~\cite{GirdharIROS2011, pizarro, Kemna-2018-986, Stuntz2016}. Exploitation, however, remains a challenge, because the primary way that users specify and guide the robot's navigation is by providing a sequence of waypoints that the robot must traverse. This is limiting because it places the onus of deciding exactly where to look on the user, before deployment, thus allowing for little adaptation in the field. 

\begin{figure}[t]
\includegraphics[width=\linewidth]{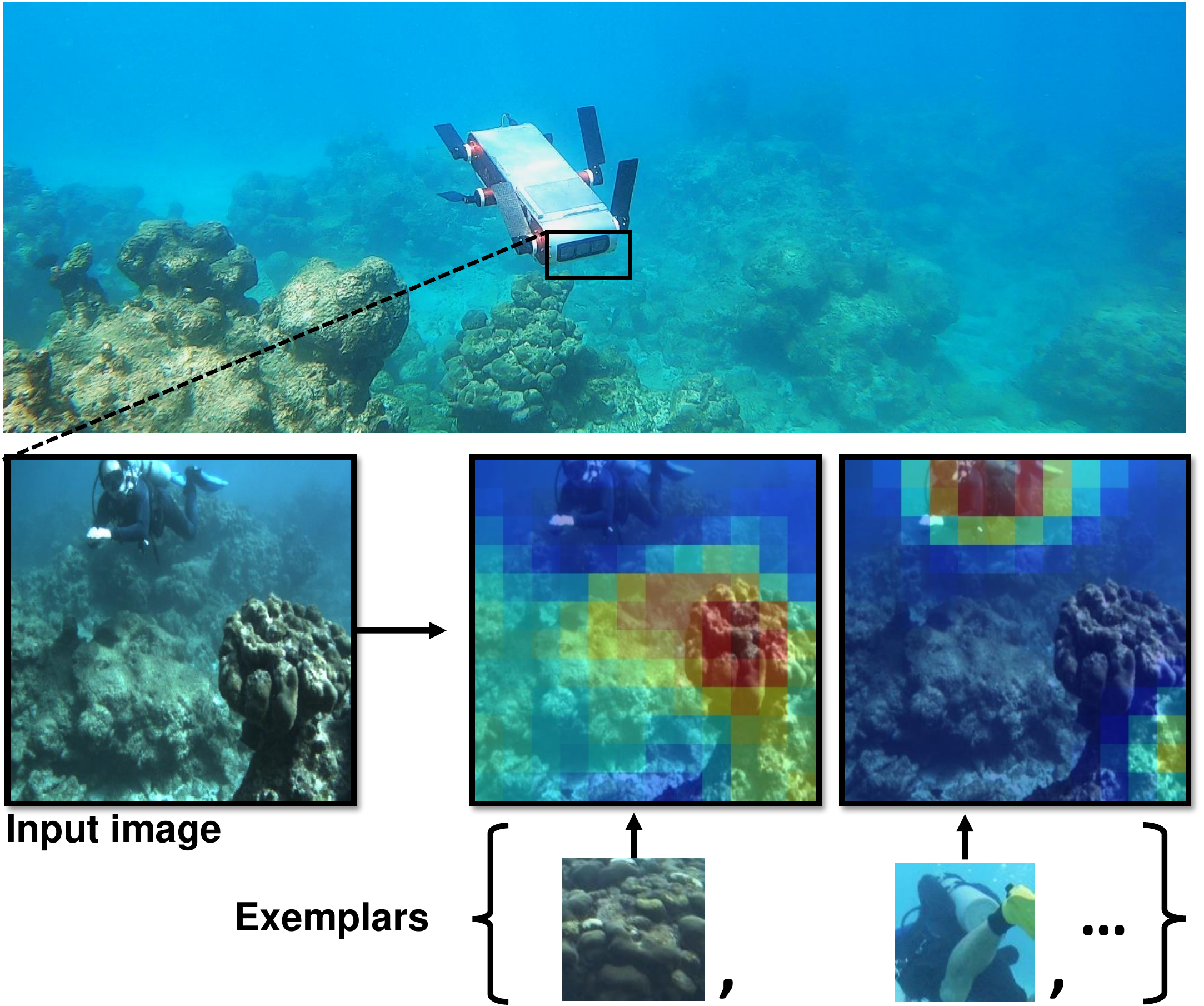}
\caption{The output of our visual similarity operator, which is informed by a single exemplar image, given by the user. The exemplar image dictates the behavior of the similarity operator. Note that the exemplars are not objects in the input image; they are objects from the same class, collected from different viewpoints and appearances.}
\label{fig:beauty}
\vspace{-5pt}
\end{figure}

In this paper, we propose a weakly supervised method for learning a visual similarity operator, whose output is shown in Fig.~\ref{fig:beauty}. This similarity operator enables informed visual navigation and allows the robot to focus its camera on parts of the scene that the user might deem relevant. Given a set of one or more exemplar images provided by the user, our method finds similar parts of the scene in the robot's camera view, under diverse viewpoints and appearances. The resulting similarity heatmap is used to guide the robot's navigation behavior to capture the most relevant parts of the environment, while performing auxiliary navigation tasks, such as visual tracking or obstacle avoidance.

\begin{figure*}[t!]
\centering
\includegraphics[width=1\textwidth]{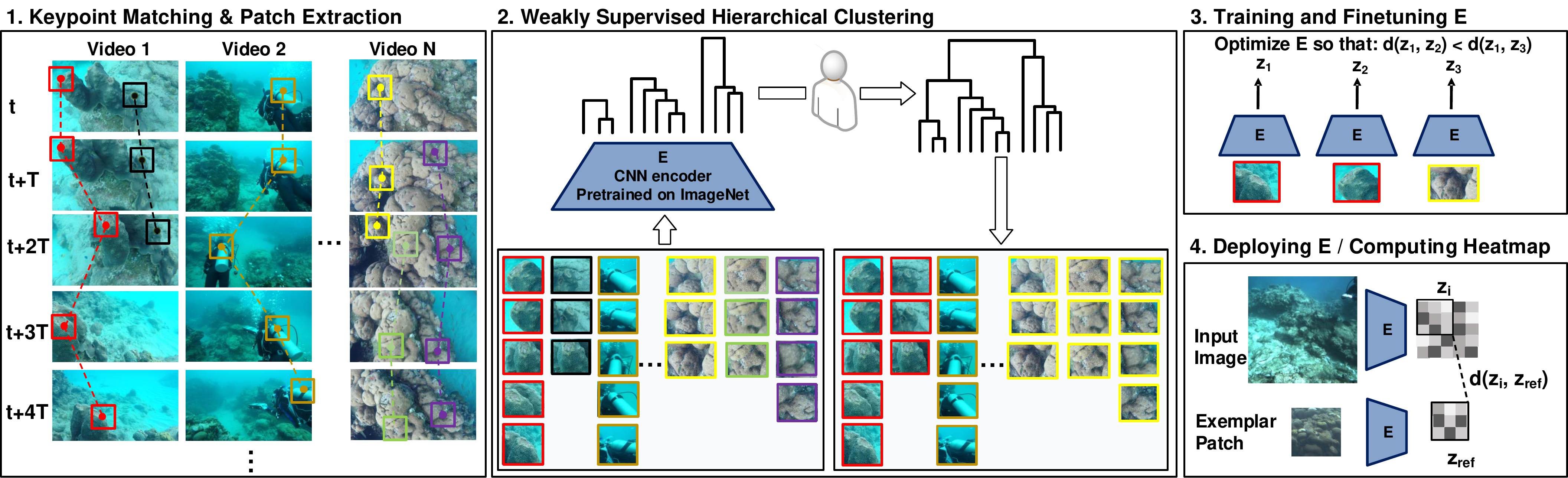}
\caption{Overview of our visual similarity training and deployment pipeline.}
\label{fig:similarity_op}
\vspace{-0.5cm}
\end{figure*}

Our method relies on representation learning from videos, and leverages traditional keypoint tracking, which yields patch sequences that locally capture the same part of the scene potentially from multiple viewpoints, as the camera is moving. We then optimize for representations that cluster multiple patch sequences, so that similar sequences are nearby in representation space, while non-similar sequences are pushed apart, via the triplet loss (see Fig.~\ref{fig:similarity_op}).

The main contributions of our paper are threefold: (a) We show that visual similarity operators can be trained to account for multiple viewpoints and appearances, not just identify a single object known to exist in the scene. (b) While fully supervised learning techniques for video often require at least a full day of annotations, our method requires about two hours of weak supervision, which is practical for field deployment. (c) While most robotic visual search methods are deployed indoors in 2D, or in constrained environments, we demonstrate visual search in 3D, in the open ocean, over hundreds of cubic meters of working volume.

\section{RELATED WORK}
\label{sec:related}
\textbf{Image Retrieval and Reverse Image Search:} Existing literature on unsupervised representation learning for computer vision is broad. We choose to focus our discussion on similarity learning for visual search and object localization in an image, which has traditionally been a key concern in image retrieval~\cite{Radenovic_2018_CVPR} and reverse image search for search engines, and which has been relying on embeddings from convolutional networks~\cite{Babenko_2014_eccv_neural_codes, GordoARL_2016_eccv, Gordo_2016_IJCV, jun2019combination, Revaud_2019_ICCV}. Aside from image retrieval, unsupervised object discovery and localization~\cite{Cho_2015_CVPR, Coates_2012_lncs, He_2018_DOAP, doersch_2015_ICCV, icml/BojanowskiJ17, Vo_2019_CVPR, Paulin_2015_ICCV, deep_clustering_fair, He_2019_CoRR} has also made use of learned visual representations, trained on auxiliary tasks. These tasks typically include ImageNet classification, unsupervised reconstruction-based losses, and egomotion-based prediction for feature extraction~\cite{AgrawalCM_2015_ICCV, Pathak_2017_CVPR, object_discovery_fox, gupta_videos}. Our method is most similar to~\cite{gupta_videos}, however, we use Weakly supervised hierarchical clustering to increase the quality of selecting similar vs dissimilar examples.      

\textbf{Visual Attention Models:} Computational models of visual attention~\cite{730558} in robotics promise to increase the efficiency of visual search and visual exploration by enabling robots to focus their field of view and sensory stream towards parts of the scene that are informative according to some definition~\cite{frintrop_2010}. So called \emph{bottom-up} attention methods define an image region as salient if it is significantly different compared to its surroundings or from natural statistics. Some of the basic features for bottom-up attention have included intensity gradient, shading, glossiness, color, motion~\cite{frintrop_2010, Tsotsos_2011, Li_2014} as well as mutual information~\cite{Bruce_2008, GOTTLIEB2013585}. 

The ability of bottom-up features to determine and predict fixations is not widely accepted~\cite{doi:10.1080/09541440500236661}, especially when a top-down task is specified, such as ``find all the people
in the scene''. \emph{Top-down} attention models are task oriented, with knowledge coming externally from a user that is looking for a particular object. Some of the first attention models to have 
combined these two types of attention are: Wolfe's Guided Search Model~\cite{Wolfe1994} which comprises a set of heuristics for weighing bottom-up feature maps, and the Discriminative Saliency Model~\cite{NIPS2004_2567}, which 
defines top-down cues as the feature maps that minimize the classification error of classes specified by the user. There's also the Contextual Guidance Model~\cite{1246946, Torralba06contextualguidance} 
uses the \textit{gist} descriptor~\cite{Oliva:2001:MSS:598425.598462} that provides a summary of the entire scene to guide the set of possible locations where the desired target might be. Finally, there's the Selective Tuning
Model~\cite{TSOTSOS1995507}, which relies on a hierarchical pyramid of feature maps, the top of which is biased or determined by a task and the lower levels of which are pruned according to whether they contribute to
the winner-takes-all or soft-max processes that are applied from one level of the hierarchy to the next. Notably, it does not only operate in a feedforward fashion.

\textbf{Visual Search:} We want to reward the robot for recording images that are important to the user. Combining bottom-up and top-down attention mechanisms for the purpose of getting the user the data they need is one of our objectives in this work. This has connections to existing active vision and visual search systems for particular objects~\cite{Shubina2010, zhu2017icra, zhu2017iccv, rasouli, bourgault, DICKINSON1997239, Forssen_2008_ICRA}. Top-down task specification expresses the user's evolving preferences about what kind of visual content is desired. We treat this bottom-up and top-down visual attention model as a user-tunable \emph{reward function/saliency map} for visual content that guides the robot's visual search, so that it records more footage of scenes that are deemed important by the user. We contrast \textit{visual search} with \textit{visual exploration} strategies, many of which focus the camera towards parts of the scene that are surprising with respect to a summary of the history of observations~\cite{yogesh_1, yogesh_2, yogesh_3, 6907021}.

\textbf{Underwater Navigation:} Most existing examples of underwater navigation rely on path and trajectory tracking, given waypoint sequences~\cite{Shkurti12iros, iros2014_megerShkurtiCortesPozaGiguereDudek, Meghjani:2014:ARS:2623380.2623579, GirdharIROS2011, pizarro, WILLIAMS2016158, Kemna-2018-986, Stuntz2016, Manderson2020rss}. Although this navigation strategy is closed-loop with respect to faithfully traversing the waypoints, there is no guarantee that useful data will be collected at those waypoints. The robot is unaware of its user's preferences and objectives for data collection, and it cannot adapt to the scene it encounters.

%it sees at the specified waypoints/viewpoints.     

%\textbf{More stuff:} 

%\begin{itemize}
%  \item general metric learning ~\cite{roth2020revisiting}
%  \item unsupervised segmentation, few-shot segmentation ~\cite{cvpr/ZhangLLYS19, XiaWNet2017}
%  \item non-patch based learning of descriptors ~\cite{DeToneMR_2018_CVPR}
%  \item semi-supervised learning from video: \cite{MisraSH_2015_CVPR}
%  \item coral segmentation, scott reef 25, moorea, etc: \cite{Williams_2006_Oceans, Steinberg_2010_IROS, Alonso_2019_JFR_CoralSeg, Bryson_2013_IROS, Beijbom_2012_CVPR_MooreaDataset, King_2018_CVPRWorkshop}
%  \item robotic visual search: ~\cite{Shubina2010} 
%  \item weakly supervised place recognition ~\cite{netvlad}
% \item agglomerative clustering ~\cite{Coates2012, Yang_2016_CVPR} 
%\end{itemize}

\begin{figure}[t]
\includegraphics[width=\linewidth]{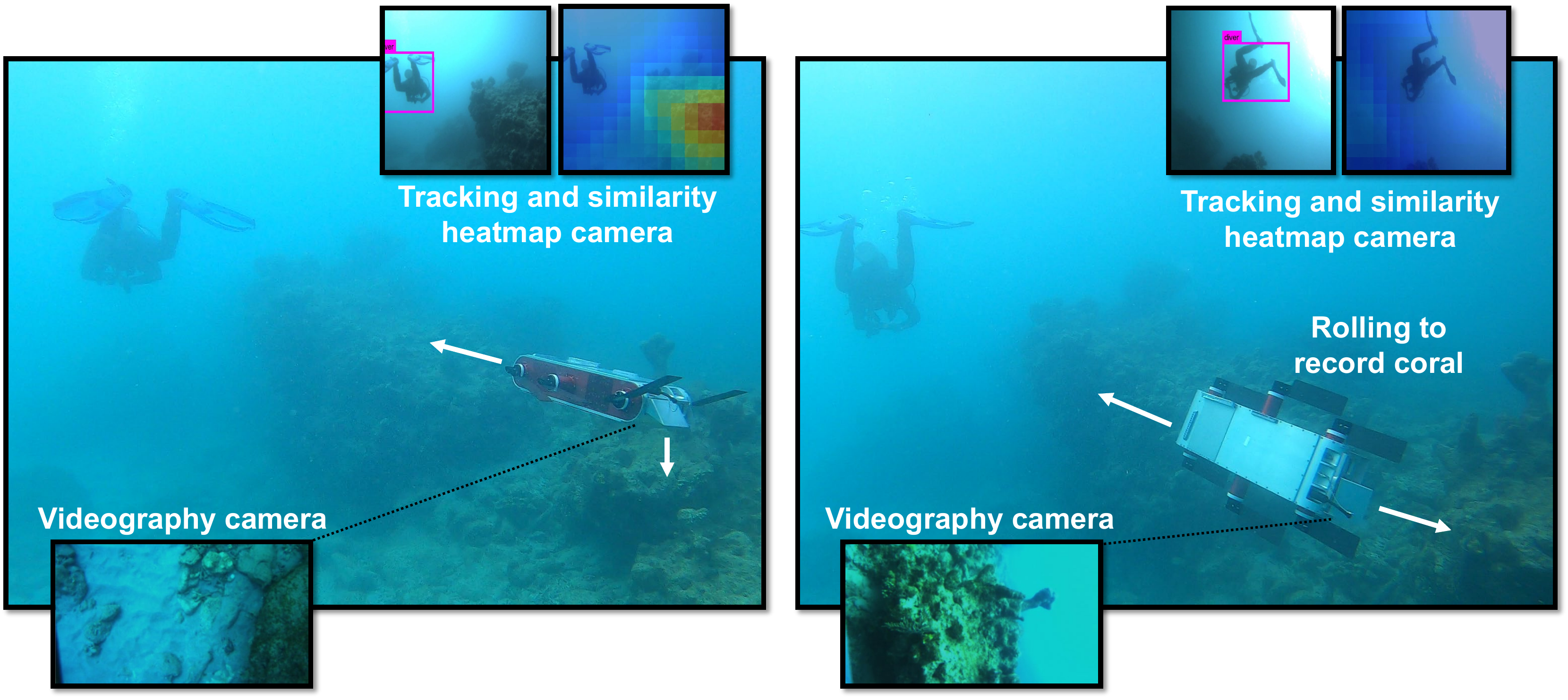}
\caption{Overview of the rolling policy employed by the robot, while diver tracking. On the left image, the robot tracks the diver and computes a heatmap based on its front cameras, while the videography camera is looking mostly at sand. The heatmap dictates rolling right as the best option. On the right image, the robot rolls in order to point its downwards-looking videography camera towards the corals, as determined by the heatmap it computed. It maintains diver tracking while doing so.}
\vspace{-10pt}
\label{fig:rolling_strategy}
\end{figure}

\section{METHODOLOGY}
\label{sec:method}
%\todo[inline]{90\% done, needs serious review for coherence}
%\todo[inline]{Left 2 write: last two subsubsections, rolling + veering off diver}

% \begin{figure}[t]
% \includegraphics[width=0.9\linewidth]{figures/kpt_tracking}
% \caption{Patch extraction from videos through keypoint tracking.}
% \label{fig:kpt_tracking}
% \end{figure}

% \begin{figure*}[t]
% \centering
% \includegraphics[width=0.8\textwidth]{figures/clustering_and_training2}
% \caption{Overview of our visual similarity training and deployment pipeline. Step 1 shown in Fig.~\ref{fig:kpt_tracking}. }
% \label{fig:similarity_op}
% \end{figure*}

Our goal is to learn a visual similarity operator which will enable more informed visual navigation for underwater robots deployed in the field. In the following sections, we discuss the various subsystems that enable our navigation strategy. In section~\ref{sec:method:learning_from_video} we present how we leverage large sets of unlabeled underwater videos to automatically extract sequences of patches built from tracked keypoints, which avoids the need for labor-intensive manual semantic annotations. In section~\ref{sec:method:learning:triplet_learning}, we then discuss training a triplet network by ranking triplets built from clusters of the tracked patch sequences to learn a similarity network. Finally, in section~\ref{sec:method:learning:heatmaps}, we present how the similarity network can be used to build attention heatmaps using exemplars provided by an operator to inform the robot's navigation. An overview of the system is shown in Fig.~\ref{fig:similarity_op}.

\subsection{Representation Learning from Videos}
\label{sec:method:learning_from_video}

Due to the low availability of semantically annotated underwater datasets, we opt to rely on unlabeled video datasets. In videos, temporal context can be leveraged as a supervisory signal as object instances will likely appear for more than one frame at a time. Furthermore, consecutive scenes with similar object instances will be available from varied viewpoints and distances, which will enable the learning of more viewpoint-invariant representations.

\subsubsection{Keypoint matching and patch extraction}
\label{sec:method:learning:feat_ext}
\label{sec:method:learning:patch_tracking}
In order to obtain semantically similar image patches from videos, we use Oriented FAST and rotated BRIEF (ORB)~\cite{Rublee_2011_ORB}, a fast and robust local appearance-based feature descriptor that allows us to quickly extract similar keypoints across frames. We extract 128 by 128 patches around ORB-extracted keypoints at each frame and track them across the video. 

We track our candidate keypoints by brute force matching the descriptors in order to find the best match across consecutive frames and we opt to only keep the top 20 best matches that satisfy a matching distance threshold. We then save the patches extracted around each tracked keypoint as a tracked patch sequence. This tracked patch sequence will provide the semantic similarity cues that we will leverage when training our triplet network. Depending on the camera's movements and speed, frame rate adjustments of the video were applied so relevant keypoints can be tracked fast enough, as well as to extract patch sequences that had significant variations and to avoid very redundant patches (as in a slow moving video). 

\label{sec:method:learning:rand_patch_ext}
Certain texture-less areas (such as sand or water) are difficult to track using traditional local descriptors, so we instead opt to randomly extract patches from videos mostly containing such areas.

\subsubsection{Weakly supervised hierarchical clustering}
\label{sec:method:learning:unsup_clust}

% \textbf{Unsupervised clustering: }
The tracked patch sequences capture very localized notions of similarity. In order to better generalize across environments and viewpoints, we use agglomerative clustering~\cite{Ward_1963_AggClust} to merge tracked patch sequences that are similar. We use the ResNet18~\cite{ResNet} model pre-trained on ImageNet and extract the last convolution layer embeddings (conv5) of a randomly sampled patch from each patch sequence and proceed with the clustering. We then build the clusters using the first and last patch in a sequence based on where the randomly sampled patch was clustered. Agglomerative clustering is a bottom-up hierarchical clustering approach, where each patch starts as a cluster and patches get merged as we move up the cluster hierarchy. We over-cluster our patch sequences, merging only the most similar patch sequences in the process.

% \textbf{Weakly supervised clustering: }
Now that the number of clusters is manageable, a human annotator can quickly manually merge clusters. While optional, as shown in section~\ref{sec:evaluation}, this step provides an additional boost in performance while requiring orders of magnitude less human annotation efforts. After merging, we are left with $k$ clusters forming the set of clusters $C$.

\subsubsection{Training and fine-tuning}
\label{sec:method:learning:triplet_learning}

We build a triplet network~\cite{Hoffer_2015_TripletNet} using three instances of the ResNet18~\cite{ResNet} architecture neural network as the baseline encoding framework and share parameters across the instances. We use the 18 layers variant of ResNet and sacrifice the accuracy usually gained from using a large variant in favor of runtime speed. We use the ImageNet~\cite{imagenet} pre-trained weights as the initialization of the network, discard the fully-connected layers after the \textit{conv5} layer. We add an $L_2$ normalization layer and flatten the output. Finally, we fine-tune the network using a triplet loss~\cite{Schroff_2015_Facenet, Hermans2017_TripletLoss} which minimizes distance between an anchor and positive patch while maximizing the distance between the anchor and a negative patch. 

Using the clusters formed in section~\ref{sec:method:learning_from_video} we build triplets of patches, $(z_A, z_P, z_N)$, formed by sampling from the set of clusters $C$ with the following conditions: for $C_k \in C$, $(z_A, z_P) \in C_k$
and $z_N \notin C_k$. The triplet loss we use is defined as follows:
\begin{equation}
    L(z_A,z_P,z_N) = \text{max}(0, m + d(z_A, z_P) - d(z_A, z_N))
\end{equation}

\noindent where $d$ is the cosine similarity between the output flattened descriptors. The margin $m$ enforces a minimum margin between the $d(z_A, z_P)$ and $d(z_A, z_N)$ and is set to $0.5$.

We train a triplet network as opposed to its siamese (pairwise loss) counterpart, which would have relied on pairs of examples, along with a similar/dissimilar label. Using a pairwise contrastive loss implies a need for labels which are contextual, and loses the relative information presented by a triplet. With a pairwise loss, two coral patches of the same species may be considered similar when the dataset includes a large variety in coral species, but dissimilar if the dataset largely consists of the same species and a more refined understanding of similarity is required. On the other hand, using the triplet loss allows us to build triplets where all three coral patches could belong to the same species but are ranked according to how similar their species-specific characteristics are. Given that our training data leverages feature-tracked sequences, we have access to more rich notions of similarity this way.

\textbf{Implementation Details:} We train using the Adam optimizer \cite{DBLP:journals/corr/KingmaB14} with a learning rate of 0.0006. During each training step, we perform semi-hard triplet mining~\cite{Schroff_2015_Facenet} in order to ensure that batches don't exclusively include easy triplets (where $d(A, N)$ is much larger than $d(A, P) + m$). Another limitation is that the number of candidate training triplets grows cubically as the dataset increases. Therefore, we have to sample a subset of the triplets that can be formed across the different clusters. We ensure that a roughly equal distribution of triplets are formed from each cluster. Once trained, the network can be used to extract a $(4, 4, 512)$ descriptor for an RGB patch of dimensions $(128, 128, 3)$.

\subsubsection{Deployment and computing heatmaps}
\label{sec:method:learning:heatmaps}
%\todo[inline]{clean up description of descriptors and their dimensions, either commit to math symbols or stick to english}
%\todo[inline]{clean up sliding descriptor discussion}

Once learned, the visual similarity network can be deployed in the field to enable more informed robotic navigation by encoding rankings of what a scientist deems important and finding related content. To do this, the operator provides a set of exemplar patches containing visual subjects of interest (in our case, particular species of coral). Given an input image and $n$ exemplars, we build a heatmap $H_k$ for each exemplar ${z_\text{ref}}_k$ by doing the following:
\begin{itemize}
    \item Extract the descriptors of the input image and the exemplar forward passes through the similarity network. Since the input image is larger $(512, 512, 3)$ than the patch image, we get a descriptor of dimensions $(16, 16, 512)$.
    \item Slide the exemplar patch descriptor over the input image's descriptor, flatten and normalize both descriptors and compute a dot product matrix.
    \item Upsample the computed similarity matrix to the original image's dimensions to form our heatmap.
\end{itemize}

\noindent An overview of the heatmap construction is presented in part 4 of Fig.~\ref{fig:similarity_op}. Example heatmaps computed from two different exemplars are shown in Fig.~\ref{fig:beauty}. We then generate a final weighted heatmap by merging the individual heatmaps along with an optionally-provided ranking of the exemplars, in terms of attention priority. This final heatmap acts as an attention model for the robot to follow as it navigates, prioritizing areas that are most similar to the exemplars provided by the operator. 

\subsection{Informed Visual Search While Diver Tracking}

The first way in which we deploy our visual similarity operator on an underwater 
robot is in the context of visual navigation by tracking a diver. This avoids the need for mapping and localization, and simplifies the navigation process considerably. In this case the robot uses its front cameras to track the diver and compute a heatmap. It uses its back camera, which is downward-looking, as the main videography camera recording in high resolution. This is shown in Fig.~\ref{fig:rolling_strategy}.

For each incoming front camera frame (at about $10$Hz) the heatmap is divided into three regions of equal area. The one that has the highest cumulative score determines the rolling direction of the robot for a fixed number of seconds (in our case, $10s$). The rolling direction of the robot is in $[-90, +90]$ deg, and the diver tracking controller~\cite{convoying_iros_2017} is still running. After rolling, the robot returns to its flat orientation to compute the heatmap once again.

\begin{figure}[t]
\includegraphics[width=\linewidth]{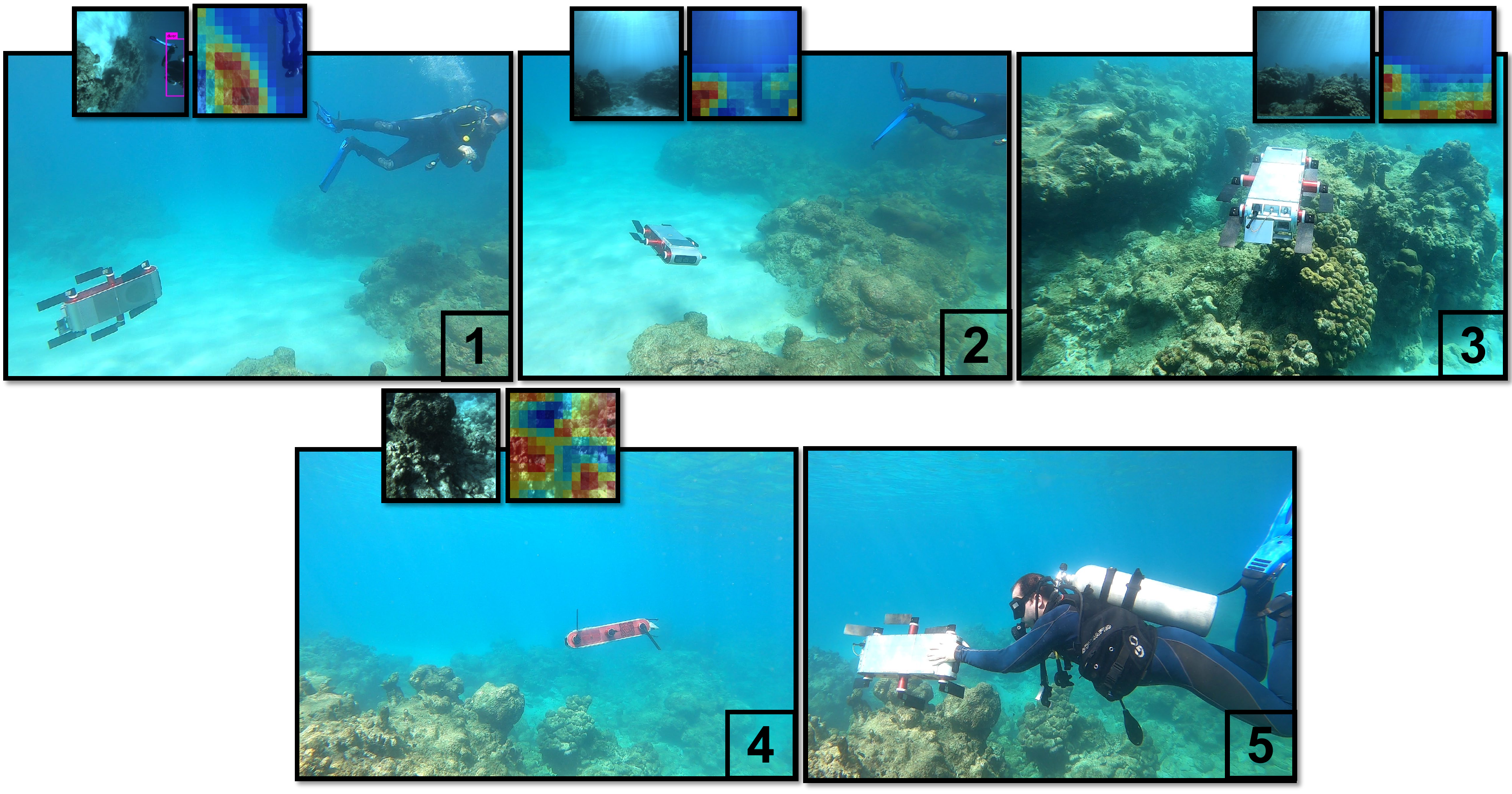}
\caption{Overview of the full field experiment run in the Caribbean Sea off the west coast of Barbados. Footage of this experiment can be found in the accompanying video. Summary: (1) While diver tracking, the robot finds areas of interest as determined by the heatmap and rolls to point its back camera towards them, for better recording. (2) The robot detaches from the diver and autonomously exploits the area, after detecting a part of the scene that is highly relevant. (3) The robot avoids obstacles based on the visual navigation method in~\cite{Manderson2018iros} and collects close up footage of the reef. (4) The robot is done exploring when it sees something relevant, about which it decides to alert the diver, and circles around the location of interest until the diver notices (5) The diver manually rolls the robot to 45 deg to signal that he saw the location of interest the robot had identified, and that the robot can continue searching.}
\label{fig:diver_detach}
\end{figure}

\begin{figure*}[t]
\centering
\includegraphics[width=0.9\linewidth]{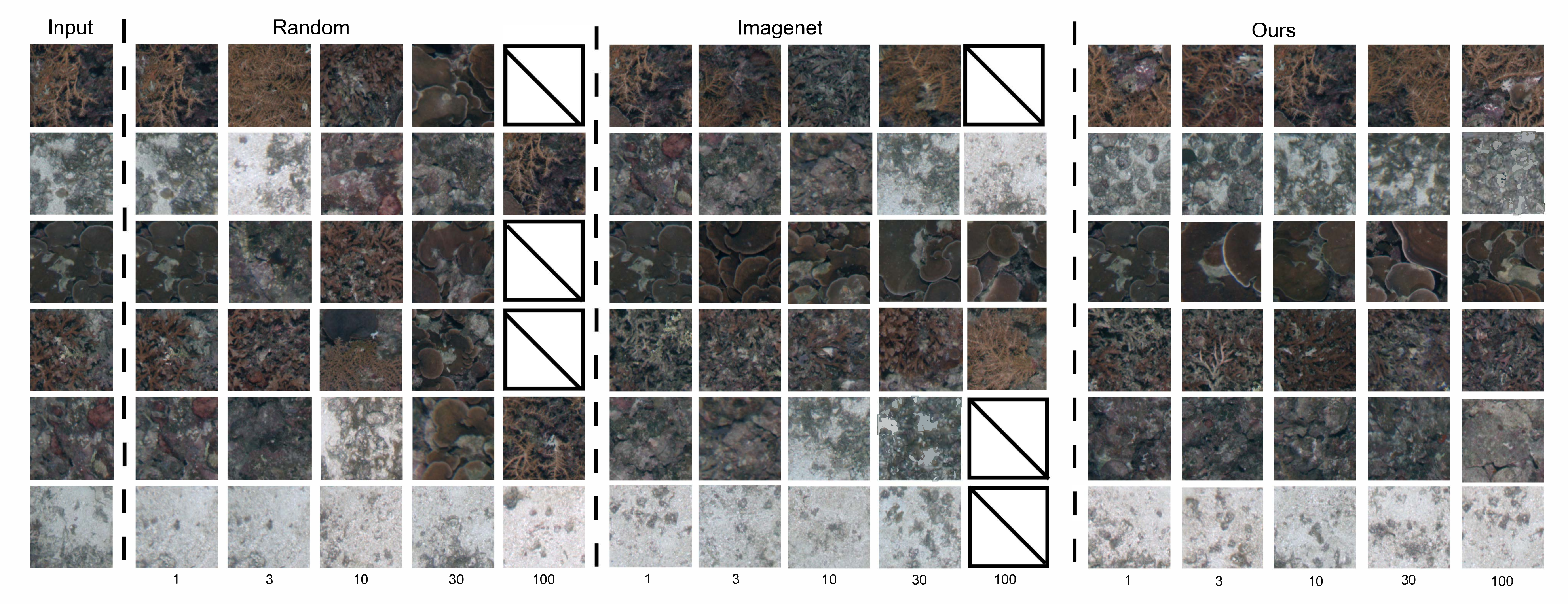}
\vspace{-15pt}
\caption{Example of patches and their ordering in the cluster on \textit{Scott Reef 25}. The exemplar is shown on the left. We then show the 1st, 3rd, 10th, 30th, and 100th retrieved patch. We use our Weakly supervised method and compare with a randomly initialized ResNet18 and pre-trained on ImageNet. Blank striked boxes indicate that the number of retrieved patches were less than that particular rank.}
\label{fig:cluster_qualitative}
\vspace{-7pt}
\end{figure*}

\subsection{Informed Visual Search and Autonomous Visual Navigation}
The second way in which we deploy our similarity operator on an underwater 
robot is in the context of autonomous visual navigation that combines obstacle avoidance and relevant-data-seeking behavior. This is shown in Fig.~\ref{fig:diver_detach}. 

After the robot identifies a relevant part of the scene during its rolling motion,
it decides whether to keep following the diver or detach and start searching on its own. We make this decision based on the heatmap values and coverage in the image. 
When the robot stops following the diver, it needs to avoid obstacles. We do this via the visual navigation method in~\cite{Manderson2018iros}, based on our existing work. 
This method learns a vision-based Bayesian Neural Network policy via fully-supervised imitation learning. This policy is a spatial classifier that outputs the desired direction for pitch and yaw angles of the robot, which are then tracked by the low-level autopilot~\cite{iros2014_megerShkurtiCortesPozaGiguereDudek}.    

The robot keeps searching, using the strategy above, until it sees something highly relevant. In our setup, the robot then needs to alert the diver about the location of interest that it found. Given that we do not have a sufficiently high-power sound source on the robot, we do this by making the robot swim in circles around the location of interest, maintaining depth, until the diver notices. When the diver approaches the robot, it manually rolls the robot to 45 deg, signalling that he has seen the location of interest the robot identified, and that the robot can continue its search, either by tracking the diver or by autonomous navigation.

% \begin{figure*}[b]
% \begin{minipage}{0.5\columnwidth}
\begin{table*}[t]
\centering
\begin{tabular}{c | c | c c c c | c c c c} 
 \multicolumn{2}{c}{} & \multicolumn{4}{c}{\textbf{Scott Reef 25}} &  \multicolumn{4}{c}{\textbf{Bellairs Reef}} \\
 \hline
 Model name & Exemplars & Acc & P & R & F1 & Acc & P & R & F1 \\ [0.5ex] 
 \hline
 CVAE (Weakly supervised) & 1 & 0.40 & 0.38 & 0.40 & 0.35 & 0.31 & 0.37 & 0.31 & 0.28 \\
 CVAE (Weakly supervised)  & 5 & 0.40 & 0.39 & 0.40 & 0.31 & 0.31 & 0.27 & 0.31 & 0.25 \\
 CVAE (Weakly supervised)  & 10 & 0.43 & 0.57 & 0.43 & 0.32 & 0.34 & 0.42 & 0.34 & 0.30 \\
 IIC (Unsupervised) & NA & 0.28 & 0.14 & 0.28 & 0.18 & 0.33 & 0.16 & 0.33 & 0.21 \\
 Randomly Initialized & 1 & 0.24 & 0.43 & 0.24 & 0.13 & 0.32 & 0.43 & 0.32 & 0.30 \\
 Randomly Initialized & 5 & 0.21 & 0.21 & 0.21 & 0.08 & 0.31 & 0.25 & 0.31 & 0.22 \\
 Randomly Initialized & 10 & 0.21 & 0.15 & 0.21 & 0.07 & 0.33 & 0.26 & 0.33 & 0.26 \\
 Pre-trained (ImageNet) & 1 & 0.68 & 0.74 & 0.68 & 0.65 & 0.67 & 0.70 & 0.67 & 0.68 \\
 Pre-trained (ImageNet) & 5 & 0.81 & 0.83 & 0.81 & 0.81 & 0.73 & 0.77 & 0.73 & 0.71 \\
 Pre-trained (ImageNet) & 10 & 0.82 & 0.86 & 0.82 & 0.81 & 0.71 & 0.76 & 0.71 & 0.70 \\
 Unsupervised hierarchical clustering & 1 & 0.90 & 0.91 & 0.90 & 0.90 & 0.62 & 0.72 & 0.62 & 0.62 \\
 Unsupervised hierarchical clustering & 5 & 0.91 & 0.92 & 0.91 & 0.91 & 0.69 & 0.74 & 0.69 & 0.68 \\
 Unsupervised hierarchical clustering & 10 & 0.91 & 0.92 & 0.91 & 0.91 & 0.66 & 0.74 & 0.66 & 0.66 \\
 Weakly supervised hierarchical clustering (ours) & 1 & \textbf{0.97} & \textbf{0.97} & \textbf{0.97} & \textbf{0.97} & \textbf{0.78} & \textbf{0.80} & \textbf{0.78} & \textbf{0.77} \\
 Weakly supervised hierarchical clustering (ours) & 5 & \textbf{0.97} & \textbf{0.97} & \textbf{0.97} & \textbf{0.97} & 0.77 & 0.79 & 0.77 & 0.77 \\
 Weakly supervised hierarchical clustering (ours) & 10 & \textbf{0.97} & \textbf{0.97} & \textbf{0.97} & \textbf{0.97} & 0.77 & 0.79 & 0.77 & 0.77 \\
 \hline
\end{tabular}
\caption{Classification evaluation}
\label{table:classification_results}
\vspace{-20pt}
\end{table*}
% \end{minipage}
% \end{figure*}

\section{EVALUATION}
\label{sec:evaluation}

%\todo[inline]{intro is a lil rough}
We evaluate our system on two underwater datasets as well as in the field. We measure our similarity network's ability to classify patches when shown exemplars from each class. We also evaluate our system's ability to generate semantic segmentation by leveraging the heatmaps given exemplar patches from the relevant classes. Finally, we deploy our system on the Aqua robot, an amphibious hexapod robot \cite{sattar2008enabling}, in an underwater environment and record its ability to collect more relevant images using a camera pointing policy.

\subsection{Underwater Datasets}
\label{sec:evaluation:datasets}
We perform our evaluations on two underwater datasets for which we train the system separately: 

\subsubsection{Bellairs Reef dataset} 
\label{sec:evaluation:datasets:bbd}
a collection of 56 videos captured with GoPro and Aqua on-board cameras at different sites in the Caribbean sea off the West coast of Barbados near the McGill Bellairs Research Institute. The videos feature a variety of scenery, viewpoints, and lighting and turbidity conditions. Mainly, they cover footage of sand areas, various dead and live coral and diver-robot activity. The dominant class labels that appear are 1. Sand, 2. Diver, 3. Aqua robot, 4. Dead coral, 5. Finger-like corals (porites, etc), and 6. Spheroid-shape stony corals (brain, starlet, dome, etc). We split the dataset into 38 videos used for training and 18 for evaluation. The videos are 5 to 15 minutes long. In general, patch sequences tracked from this dataset span longer timespans as the camera moves slowly, at the pace of an exploring diver. The mean patch sequence length is 4 tracked patches, while the longest sequence spans 30 patches. 

\subsubsection{Scott Reef 25 dataset} 
\label{sec:evaluation:datasets:sr25}
exclusively top-down stereo camera imagery collected with an AUV at Scott Reef in Western Australia over a 50 by 75 meter full-coverage of the benthos \cite{Steinberg_2010_IROS, Bryson_2013_IROS}. We use 9831 RGB images captured by the left camera. The images cover areas of dense coral (multiple species), sand and transition areas in between. The dominant class labels that appear in the data are 1. Sand, 2. Mixture, 3. Finger-like coral (isopora), 4. Thin birdsnest coral, 5. Table coral (acropora), and 6. Dead coral. We split the dataset temporally into 7061 training images and 2770 test images. Patch sequences tracked from this dataset have 2 patches only on average, mainly due to the high speed of the AUV. The longest sequence spans 23 patches. 

% \vspace{-9pt}
\begin{figure}[H]
\centering
\includegraphics[width=0.65\linewidth]{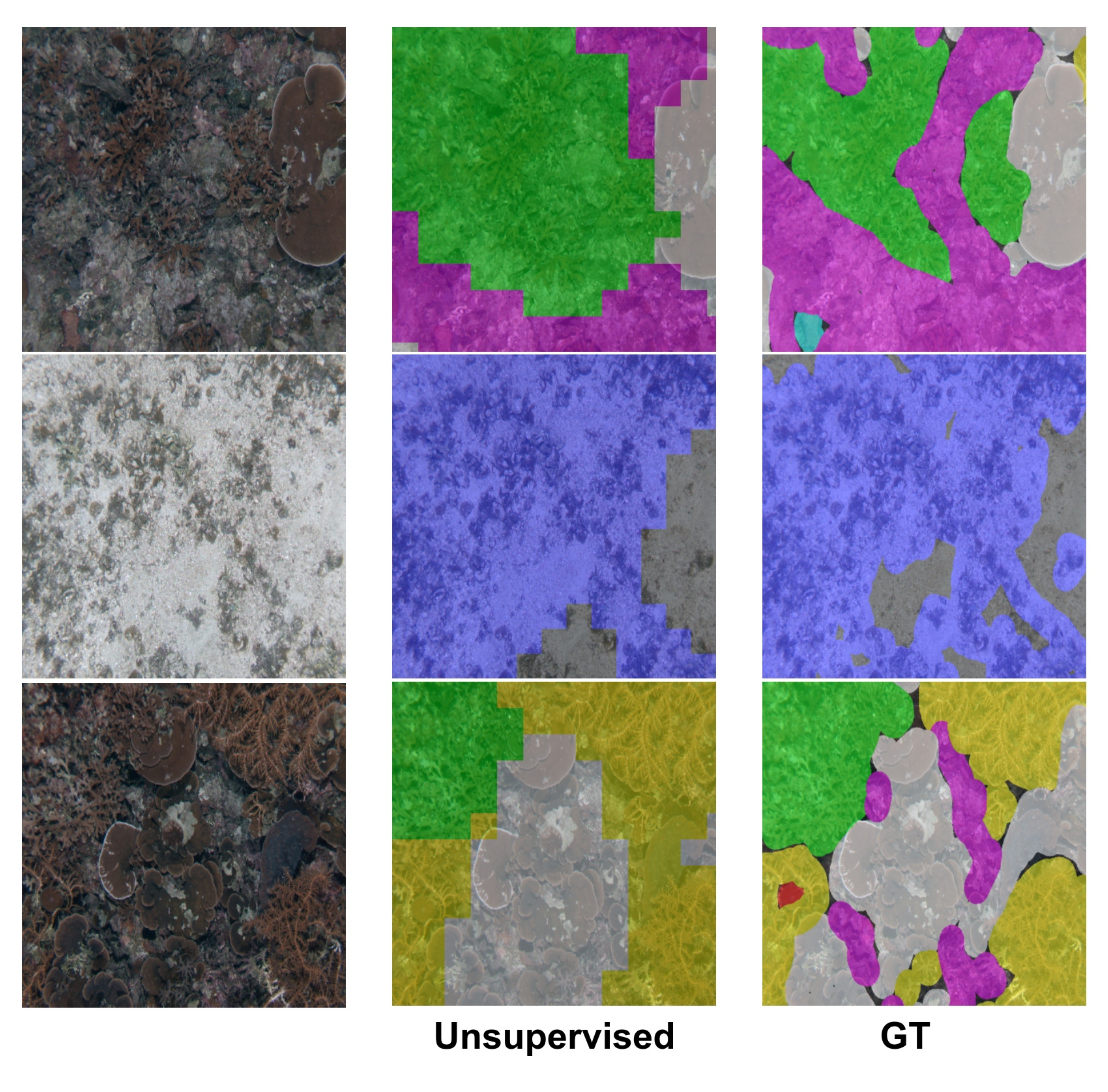}
\vspace{-10pt}
\caption{Example segmentation results on \textit{Scott Reef 25} using our unsupervised model. From left to right: input image, our predicted semantic segmentation, ground truth segmentation (GT).}
\label{fig:segm_results}
\end{figure}

\subsection{Representation Learning on Underwater Datasets}
\label{sec:evaluation:learning}

For each dataset, we train our network on the automatically clustered and Weakly supervised versions of the dataset as described in section~\ref{sec:method:learning:unsup_clust}. We now evaluate our system on classification and semantic segmentation using standard metrics.

% \vspace{5pt}
\subsubsection{Classification}
\label{sec:evaluation:learning:class}
We manually curate a subset of 1500 patches from each dataset into classification datasets labelled using the dominant classes listed in section~\ref{sec:evaluation:datasets:bbd} and section~\ref{sec:evaluation:datasets:sr25}. We then evaluate the system on the classification task by comparing the test set patches with 1, 5, and 10 randomly sampled exemplars of each class and assigning a prediction label to the class with the highest mean similarity. We summarize the classification results in Tab.~\ref{table:classification_results}. We show that the model demonstrates a boost in performance when fine-tuned on the clustered patch sequences on both datasets. As expected, by additionally merging clusters manually, we further increase the performance of the system. Note that by comparing against a higher number of exemplars per class, there is no noticeable difference, demonstrating a mostly stable representation of the classes. Only needing a single exemplar is particularly beneficial when deploying on robots with compute limitations. We compare our system against a Conditional VAE (CVAE) \cite{NIPS2015_CVAE} and Invariant Information Clustering (IIC) ~\cite{ji2018invariant} trained on both datasets and outperform them by a significant margin. As a disclaimer, IIC uses an uninitialized ResNet50 architecture while we fine-tune on an ImageNet pre-trained ResNet18 network, thus benefitting from the pre-trained weights. We used the TensorFlow implementation of IIC\footnote[1]{Github repo: \url{https://github.com/nathanin/IIC}}.

An important note is that the system is able to obtain an ordering of the retrieved patches based on their similarity to the provided exemplar as opposed to simply classifying it. This is especially important when a operator is interested in retrieving images of an object that are not as common as the class center. Example ordered retrieval results are presented in Fig.~\ref{fig:cluster_qualitative}.

% \vspace{5pt}
\subsubsection{Semantic segmentation}
\label{sec:evaluation:learning:segm}
We manually annotate pixel-level semantic segmentation masks for 250 randomly sampled images from each dataset using their respective dominant classes. To build a semantic segmentation prediction, we merge the heatmaps generated for each class by assigning every pixel to the class with the highest similarity. We summarize the results in Table~\ref{table:segm_results} and show example segmentations in Fig.~\ref{fig:segm_results}. Note that IoU is intersection over union.

\begin{table}[H]
\centering
\begin{tabular}{l | l c c c } 
 \hline
 Dataset & Model & Mean & Mean & Weighted \\
  &  & Acc & IoU & IoU \\
 [0.5ex]
 \hline
 Scott \\ Reef 25 & Randomly Initialized & 0.30 & 0.03 & 0.02 \\ 
 & Pre-trained (ImageNet) & 0.39 & 0.28 & 0.51 \\
 & Unsupervised & 0.41 & 0.31 & 0.57 \\
 & Weakly supervised & 0.42 & 0.29 & 0.52 \\ [1ex]
 \hline
 Bellairs \\ Reef & Randomly Initialized & 0.32 & 0.20 & 0.41 \\ 
 & Pre-trained (ImageNet) & 0.59 & 0.40 & 0.56 \\
 & Unsupervised & 0.46 & 0.34 & 0.54 \\
 & Weakly supervised & 0.51 & 0.36 & 0.54 \\ [1ex] 
 \hline
\end{tabular}
\caption{Segmentation evaluation}
\label{table:segm_results}
\vspace{-15pt}
\end{table}

% Not gonna use this cause results are so-so
% As a simple refinement method, we use superpixels \cite{Ren_2003_SuperPixels} to better adjust segmentations along boundaries. 

\subsection{Robot Field Trials}

\begin{figure}[t]
\includegraphics[width=\linewidth]{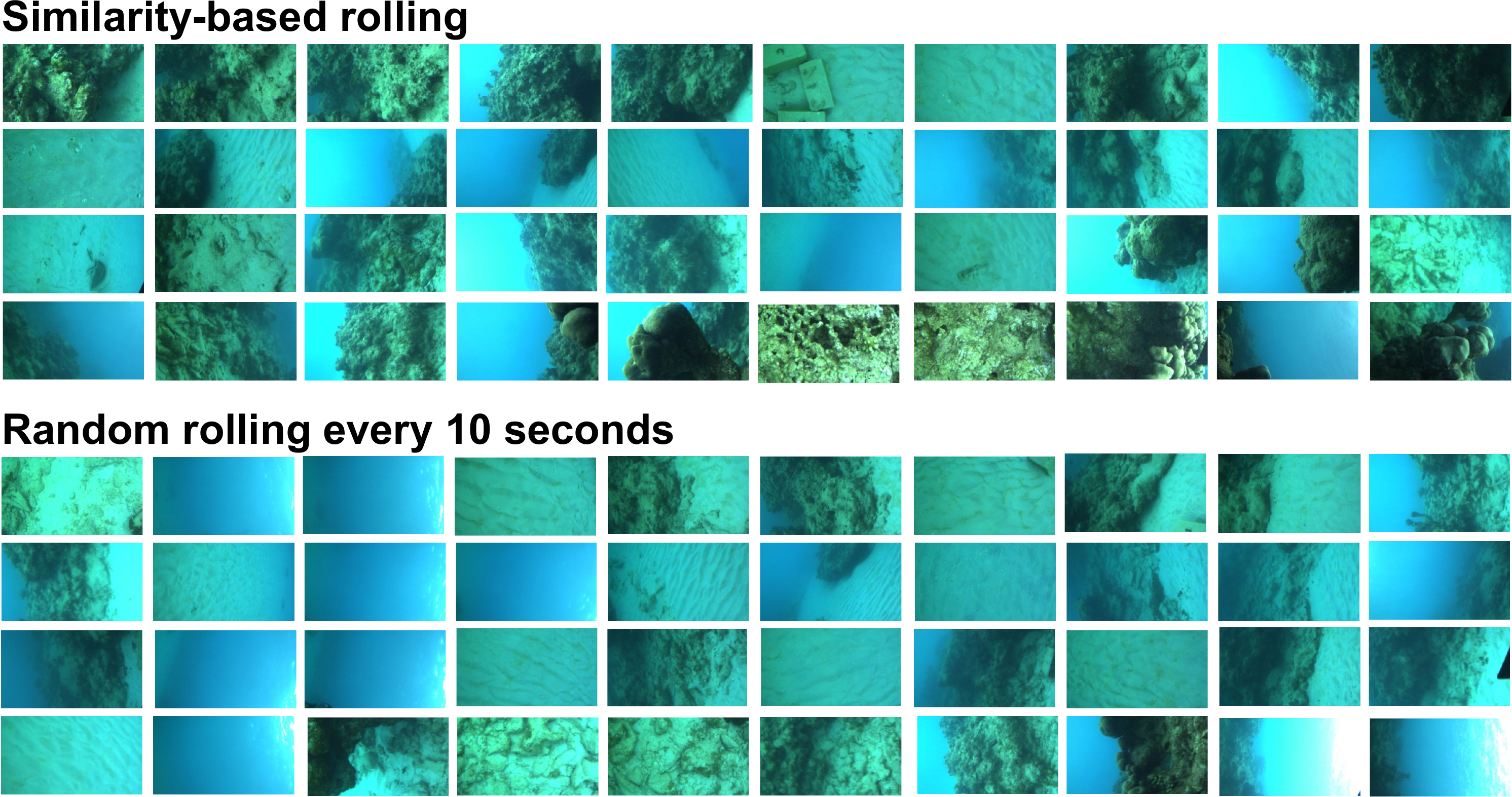}
\caption{(Top) Images collected by the back camera of the robot, while doing informed rolling based on the similarity model. 25/40 coral images recorded. (Bottom) Images collected by the same camera at the same place, when making random rolling decisions. 15/40 coral images recorded. The latter sequence contains more irrelevant images (water and sand).}
\label{fig:rolling_policy_photos}
\vspace{-10pt}
\end{figure}

We deploy our system on the Aqua underwater robot~\cite{sattar2008enabling}, in the Caribbean sea, on the west coast of Barbados, over the course of two weeks. The Robot Operating System (ROS) \cite{ROS} framework is used to handle distributed communications among Aqua’s controllers and sensor suite. We rely on the front-facing RGB camera which runs on a laptop-grade dual-core Intel NUC i3 CPU and an Nvidia GPU (Jetson TX2) \cite{Manderson_2018_Oceans_AquaGPU}. The diver tracker~\cite{convoying_iros_2017, redmon2016yolo9000} ran on the NUC at $10$Hz, while the similarity operator ran on the TX2 at $4$Hz. A representative experiment showing the improvement in collected data from executing the informed rolling policy illustrated in Fig.~\ref{fig:rolling_strategy} is shown in Fig.~\ref{fig:rolling_policy_photos}. The rolling policy informed by the similarity operator records more relevant images than the uninformed (random) rolling policy.

\section{CONCLUSIONS}

We presented a method to learn a Weakly supervised visual similarity model that enables informed robotic visual navigation in the field given exemplar images provided by an operator. Our system relies on keypoint tracking across video frames to extract patches from various viewpoints and appearances and leverages hierarchical clustering to build a dataset of similar patch clusters. By asking an operator to weakly supervise the merging of generated image clusters, we circumvent the need for tedious manual frame-by-frame annotations. We then train a triplet network on this dataset, and are able to learn useful representations which enable the computation of an attention heatmap which is used to inform the navigation of a robot underwater. We evaluate our similarity representation's classification and semantic segmentation performance on two underwater datasets and show a boost between 17\% and 32\% in retrieval and classification performance compared to simply using pre-trained on ImageNet ResNet18 embeddings. We also successfully deployed our similarity operator on the Aqua underwater robot in large-scale field trials, in which the robot and a diver/scientist collaboratively search for areas of interest and demonstrate a higher retrieval of coral images than uninformed navigation strategies. 

\addtolength{\textheight}{0cm}   % This command serves to balance the column lengths
  %                                 % on the last page of the document manually. It shortens

%%%%%%%%%%%%%%%%%%%%%%%%%%%%%%%%%%%%%%%%%%%%%%%%%%%%%%%%%%%%%%%%%%%%%%%%%%%%%%%%

%%%%%%%%%%%%%%%%%%%%%%%%%%%%%%%%%%%%%%%%%%%%%%%%%%%%%%%%%%%%%%%%%%%%%%%%%%%%%%%%

%%%%%%%%%%%%%%%%%%%%%%%%%%%%%%%%%%%%%%%%%%%%%%%%%%%%%%%%%%%%%%%%%%%%%%%%%%%%%%%%
%\section*{APPENDIX}

\section*{ACKNOWLEDGMENT}
The authors would like to acknowledge financial support from the Natural Sciences and Engineering Research Council (NSERC) of Canada.
The authors would also like to acknowledge the Australian National Research Program (NERP) Marine Biodiversity Hub for the taxonomical labeling and the Australian Centre for Field Robotics for gathering the image data in the \textit{Scott Reef 25} dataset.

\FloatBarrier
%%%%%%%%%%%%%%%%%%%%%%%%%%%%%%%%%%%%%%%%%%%%%%%%%%%%%%%%%%%%%%%%%%%%%%%%%%%%%%%%
\bibliography{bibtex/refs}{}
\bibliographystyle{bibtex/IEEEtran}

\end{document}